\pdfoutput=1

\documentclass[11pt]{article}

\usepackage{ACL2023}

\usepackage{times}
\usepackage{latexsym}

\usepackage[T1]{fontenc}

\usepackage[utf8]{inputenc}

\usepackage{microtype}

\usepackage{inconsolata}

\title{ModRWKV: Transformer Multimodality in Linear Time}

\newcommand{\method}{\textsc{ModRWKV}}
\usepackage{caption}
\usepackage{subcaption}
\usepackage{graphicx} 
\usepackage{multirow}
\usepackage{booktabs}
\usepackage{wrapfig}
\usepackage{tikz}
\usepackage{pgfplots}
\usepackage{xcolor}
\usepackage{colortbl}
\definecolor{brickred}{RGB}{203,65,84}
\definecolor{darkcyan}{RGB}{0,139,139}
\usepackage{fontawesome5}
\usepackage{amsmath, amssymb}

\author{
Jiale Kang$^{1*}$,\space Ziyin Yue$^{1*}$,\space Qingyu Yin$^{2*}$ ,\space Jiang Rui$^{1,3}$,\space Weile Li$^{1}$,\space Zening Lu$^{1}$,\space Zhouran Ji$^{1}$\\
$^1$RWKVOS, \space $^2$Zhejiang University  ,\space $^3$The Hong Kong University of Science and Technology\\
\texttt{jiale@rwkvos.com}
}

\newcommand\blfootnote[1]{%
\begingroup
\renewcommand\thefootnote{}\footnote{#1}%
\addtocounter{footnote}{-1}%
\endgroup
}

\begin{document}
\maketitle

\blfootnote{$*$ Equal contributions.}

\begin{abstract}
Currently, most multimodal studies are based on large language models (LLMs) with quadratic-complexity Transformer architectures. While linear models like RNNs enjoy low inference costs, their application has been largely limited to the text-only modality. This work explores the capabilities of modern RNN architectures in multimodal contexts. We propose ModRWKV—a decoupled multimodal framework built upon the RWKV7 architecture as its LLM backbone—which achieves multi-source information fusion through dynamically adaptable heterogeneous modality encoders. We designed the multimodal modules in ModRWKV with an extremely lightweight architecture and, through extensive experiments, identified a configuration that achieves an optimal balance between performance and computational efficiency. ModRWKV leverages the pretrained weights of the RWKV7 LLM for initialization, which significantly accelerates multimodal training. Comparative experiments with different pretrained checkpoints further demonstrate that such initialization plays a crucial role in enhancing the model's ability to understand multimodal signals. Supported by extensive experiments, we conclude that modern RNN architectures present a viable alternative to Transformers in the domain of multimodal large language models (MLLMs). Furthermore, we identify the optimal configuration of the ModRWKV architecture through systematic exploration.

https://github.com/JL-er/ModRWKV.git
\end{abstract}
\section{Introduction}
Linear complexity model~\cite{peng2025rwkv7gooseexpressivedynamic,gu2024mambalineartimesequencemodeling,yang2024gatedlinearattentiontransformers,yang2025parallelizinglineartransformersdelta, yang2024parallelizing} have emerged as an efficient alternative to the attention-based Transformer architecture~\cite{vaswani2023attentionneed, yin2024stablemask} in Large Language Models (LLMs)~\cite{touvron2023llamaopenefficientfoundation, achiam2023gpt}. Among various linear models, recurrent neural networks (RNNs)~\cite{peng2025rwkv7gooseexpressivedynamic} have become a competitive approach. Characterized by constant memory usage, RNNs can perform inference at a lower cost compared to the linearly increasing KV cache of Transformers. Recent research has also enabled their parallel training capabilities~\cite{yang2024gatedlinearattentiontransformers,yang2025parallelizinglineartransformersdelta}, facilitated by hardware-aware designs optimized for modern GPU architectures~\cite{dao2022flashattentionfastmemoryefficientexact}.

Currently, LLMs are undergoing a paradigm shift—from single-modality processing to cross-modal collaboration~\cite{liu2023visualinstructiontuning, fang2025llamaomniseamlessspeechinteraction, Chen_2022, défossez2024moshispeechtextfoundationmodel}. By leveraging transfer learning from pre-trained LLM weights, these models achieve cross-modal semantic alignment in tasks such as visual question answering and speech dialogue. However, this practice has primarily been employed within the traditional Transformer architecture. In the context of linear models, few works have expanded their understanding to modalities beyond natural language. This disparity highlights a crucial gap in the current landscape of linear models.

In this paper, we describe \method. It is the first RNN-based linear model that extends its capabilities to the cross-modal domain. \method \ is based on RWKV7, a RNN-based architecture powered by generalized delta rule with vector values gating, in-context learning rates, and relaxed value replacement rule. We hypothesize that the inherent sequential processing capabilities of RNNs, coupled with a carefully designed shared parameter base, can effectively capture both intra-modal and inter-modal dependencies across diverse data types. 

We take advantage of the RWKV7 architecture to propose an innovative unified training paradigm for multimodal fusion. \method \ adopts a lightweight shared parameter base with a modality-specific encoder framework, where simply switching the front-end encoder enables seamless transfer across multimodal tasks. This approach systematically explores the representation capabilities of RNN architectures within cross-modal semantic spaces, aiming to break the Transformer-dominated research paradigm. It offers new theoretical and practical insights into the deployment of large RNN-based models in the multimodal domain.
\begin{figure*}[ht]
    \centering

    \includegraphics[width=0.8\textwidth]{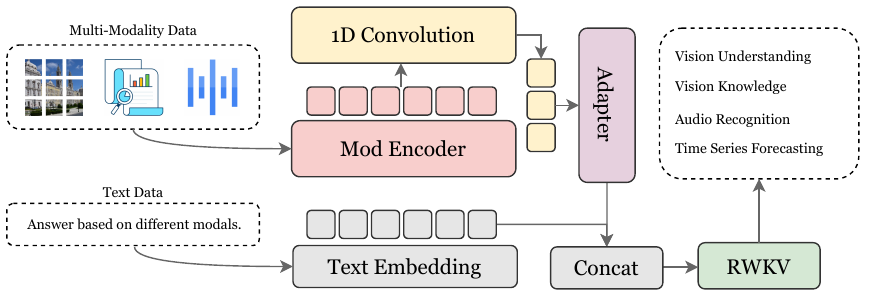}

    \caption{ModRWKV network architecture. Multi-modality data streams undergo initial processing via an encoder, a 1D Convolutional layer, and an adapter. (The 1D Convolutional layer is employed to compress the sequence length of multi-modal inputs, which significantly reduces the computational overhead during training.) Concurrently, text data is transformed through a Text Embedding module. The outputs from the adapter and Text Embedding layers are subsequently concatenated.}
    \label{fig:design}
    \vspace{-15pt}
\end{figure*}

{Our contributions can be summarized as threefold:}

\hangindent 1em
\hangafter=0
{\bf 1.} Proposed the \method \ framework, pioneering a unified multimodal training paradigm based on an RNN architecture. By adopting a plug-and-play design for modality encoders, it significantly enhances cross-modal scalability and integration efficiency.

\hangindent 1em
\hangafter=0
{\bf 2.} Conducted a comprehensive and systematic evaluation of \method's full-modality understanding capabilities, establishing a benchmark paradigm for assessing the cross-modal performance of RNN-based architectures.

\hangindent 1em
\hangafter=0
{\bf 3.} Extensive Ablation experiments validate the optimal multimodal processing design that achieves a desirable balance between performance and computational efficiency.


\section{Background}
\paragraph{RWKV7: Modern RNN Architecture} Simple linear RNNs~\cite{qiao2024vlmambaexploringstatespace, gu2024mambalineartimesequencemodeling} can be written in the following recurrent form:
\begin{equation}
    \boldsymbol{h_t} = \boldsymbol{W} \boldsymbol{h_{t-1}} + \boldsymbol{U} \boldsymbol{x_t},
\end{equation}
which enables parallelized training but lacks strong language performance and long-term dependency preservation. RWKV combines the efficiency of linear RNNs (constant memory and time complexity during inference) with powerful modeling capabilities through its time-mixing block. It uses keys $\boldsymbol{k_t}$ and values $\boldsymbol{v_t}$, linearly projected from $\boldsymbol{x_t}$, and updates the state $\boldsymbol{s_t}$ with input-dependent decay $\boldsymbol{w_t}$ and receptance $\boldsymbol{r_t}$:
\begin{equation}
    \boldsymbol{s_t} = e^{-\boldsymbol{w_t}} \cdot \boldsymbol{s_{t-1}} + \boldsymbol{k_t} \boldsymbol{v_t^T},
\end{equation}
In RWKV7, the state update is enhanced for greater expressiveness with the form:
\begin{equation}
    \boldsymbol{s_t} = \boldsymbol{G_t} \boldsymbol{s_{t-1}} + \boldsymbol{a_t} \boldsymbol{k_t} \boldsymbol{v_t^T},
\end{equation}
where employed a generalized delta rule with two improvements: (1) \textbf{In-context learning rate.} the term $\boldsymbol{a_t}$, a vector-valued learning rate projected as $\boldsymbol{a_t} = \boldsymbol{W_a} \boldsymbol{x_t}$, controls the influence of the new information $\boldsymbol{k_t} \boldsymbol{v_t^T}$ on the state. (2) \textbf{Vector value gating.} The dynamic transition matrix $\boldsymbol{G_t} = (\boldsymbol{I} - \boldsymbol{a_t} \boldsymbol{k_t} \boldsymbol{k_t^T}) \mathrm{diag}(e^{-e^{\boldsymbol{w_t}}})$ incorporates $\boldsymbol{w_t}$, a vector-valued gating parameter from $\boldsymbol{w_t} = \boldsymbol{W_w} \boldsymbol{x_t}$, enabling channel-specific decay rates. This input-dependent design makes $\boldsymbol{s_t}$ highly adaptive to context.

\paragraph{Multimodal Large Language Models}
LLMs have traditionally been trained on natural language data and are primarily designed to understand and generate text. These models excel in text-based tasks but are inherently limited to the domain of human language. Recently, many works have begun to explore the potential of large language models beyond their linguistic roots, pushing their capabilities into other modalities. From a modality perspective, MLLMs now handle a variety of data types beyond text, including images~\cite{liu2024improvedbaselinesvisualinstruction}, audio~\cite{défossez2024moshispeechtextfoundationmodel}, and video. Structurally, these models adapt by incorporating modality-specific encoders, such as visual transformers for images or audio transformers for sound. Input integration varies between unified tokenization, where all modalities are converted into a single token sequence, and cross-modal attention, where the model attends to features across modalities.


\section{Methodology}
\method \ is the first RNN-based multimodal architecture that integrates the MLLM training paradigm with a linear model, achieving exceptional hardware efficiency. In Section~\ref{sec:encoder}, we present the encoder selection design of \method. In Section~\ref{sec:adapter}, we detail the adapter design of \method. In Section~\ref{sec:seq_comp}, we describe the sequence compression method for efficiently processing diverse multimodal data.



    

\subsection{Multimodal Encoder}
\label{sec:encoder}
\paragraph{Vision Encoder.}
We evaluated CLIP \cite{radford2021learningtransferablevisualmodels} and SigLIP2 \cite{tschannen2025siglip2multilingualvisionlanguage} as alternative visual encoders for \method, applying identical adaptation frameworks to each model independently. Each vision-language encoder processes raw images to generate sequential feature embeddings that are then aligned with the RWKV large language model through lightweight adapter layers. Our experiments validated \method’s strong inherent capacity for visual information processing, with this framework demonstrating excellent cross-modal adaptability even without architectural modifications to the base language model. 

\paragraph{Audio Encoder.}
In our study, we employ WavLM \cite{Chen_2022} and Whisper \cite{radford2022robustspeechrecognitionlargescale} as audio encoders for \method. We select encoder models with sizes ranging from approximately 100M to 400M parameters, specifically choosing WavLM base+, WavLM large, Whisper small, and Whisper medium for evaluation. These encoders process audio sampled at 16,000 Hz and generate feature vectors at a frequency of 50 Hz. For the Whisper encoder, each audio segment is padded to a duration of 30 seconds.

\paragraph{Time Series Encoder.}

We adopt WaveNet~\cite{van2016wavenet} and Timer~\cite{liu2024timer} as alternative temporal encoders for \method. Timer is initialized with pre-trained weights, with the weights frozen during training, while WaveNet is trained from scratch without pre-trained weights. However, during inference, both encoders are frozen to enable zero-shot evaluation. Each encoder transforms raw time-series data into high-level feature embeddings, which are then aligned with the RWKV blocks via lightweight adapters.

\subsection{Adapter Design}
\label{sec:adapter}
We introduce a single-MLP adapter~\cite{liu2023visualinstructiontuning} for dimension alignment between modalities, reducing the adapter’s parameter. This forces the RWKV7 backbone to handle the majority of cross-modal reasoning, providing a rigorous test of RNN-based architectures in multimodal settings:
\begin{equation}
    \begin{aligned}
        \boldsymbol{h} & = \operatorname{Linear}_2(\operatorname{ReLU}(\operatorname{Linear}_1(\boldsymbol{x}))).
    \end{aligned}
    \label{eq:lora}
\end{equation}

\begin{table}[htbp]
    \centering
    \small
    \caption{Multimodal Benchmark Evaluation}
    \label{tab:multimodal_benchmarks}
    \setlength{\tabcolsep}{1pt} 
    \begin{tabular}{ll}
        \toprule
        \textbf{Benchmark} & \textbf{Description}\\
        \midrule
        VQA-v2~\cite{goyal2017vqav2} & Image Understanding\\
        TextVQA~\cite{singh2019textvqa} & Text-Image Integration\\
        GQA~\cite{hudson2019gqa} & Reasoning\\
        ScienceQA~\cite{lu2022learn} & Scientific Reasoning\\
        POPE~\cite{li2023pope} & Hallucination\\
        MMMU~\cite{yue2024mmmumassivemultidisciplinemultimodal} & Reasoning\\
        MMBench~\cite{liu2024mmbenchmultimodalmodelallaround} & Assessment\\
        LibriSpeech~\cite{7178964}& Speech Recognition\\
 Aishell-1~\cite{bu2017aishell1opensourcemandarinspeech} &Speech Recognition\\
 GIFT-Eval~\cite{aksu2024giftevalbenchmarkgeneraltime}&Time Series\\
        UTSD~\cite{liu2024timer}& Time Series\\
        \bottomrule
    \end{tabular}
    \vspace{-10pt}
\end{table}

\subsection{Sequence Compression}
\label{sec:seq_comp}
To address the computational challenges of long sequences in LLMs, we employ 1D convolution to effectively compress multimodal sequences (e.g., image patches, audio spectrograms). This approach significantly reduces processing overhead while maintaining model performance. For an input $\boldsymbol{x} \in \mathbb{R}^{C_{\text{in}} \times L}$, a convolutional kernel $\boldsymbol{W} \in \mathbb{R}^{C_{\text{out}} \times C_{\text{in}} \times k}$, stride $s \geq 1$, padding 
$p$, the $c$-th output channel $\boldsymbol{Y} \in \mathbb{R}^{C_{\text{out}} \times L'}$ is computed as:
\begin{equation}
\boldsymbol{y}_c = \underbrace{\sum_{i=1}^{C_{\text{in}}}
    \left( \sum_{j=0}^{k-1} \boldsymbol{W}_{c,i,j} \cdot \boldsymbol{x}_{i, s \cdot t + j} \right)+ b_c
}_{\text{Conv1D}}
, 
\end{equation}
where $\quad t = 0, \dots, L'-1$ and $L'$ is computed as $L' = \left\lfloor \frac{L + 2p - k}{s}\right\rfloor + 1.$
\begin{table*}[t]
    \caption{\textbf{Comparison with SoTA methods on 7 benchmarks.} 
    Benchmark names are abbreviated due to space limits. VQA-v2; GQA; SQA$^\text{I}$: ScienceQA-IMG; VQA$^\text{T}$: TextVQA; POPE; MMB: MMBench; MMMU.
    PT and IT indicate the number of samples in the pretraining and instruction tuning stages, respectively.
    }
    \label{tab:results}
    \centering
    \vspace{3pt}
    \renewcommand{\arraystretch}{1.25}
    \resizebox{\linewidth}{!}{
    \begin{tabular}{ll cc | cccc | cccc }
    \toprule
    Method & LLM  & PT & IT & VQA$^\text{v2}$ & GQA & SQA$^\text{I}$ & VQA$^\text{T}$ & POPE & MMB & MMMU \\
    \midrule
    LLaVA-1.5 & Vicuna-7B & 558K&665K&78.5&62.0&66.8&58.2&86.5&64.3&-\\
    LLaVA-1.5 & Vicuna-13B & 558K&665K&80.0&63.3&71.6&61.3&86.2&\textbf{67.7}&-\\
    LLaVA-1.6 & Vicuna-7B & 558K&665K&\textbf{81.8}&\textbf{64.2}&\textbf{72.8}&\textbf{65.7}&86.7&67.7&35.8\\
    \midrule
    LLaVA-Phi & Phi-2-2.7B & 558K& 665K& 71.4 & - &{68.4} & 48.6 & 85.0 & {59.8} & -\\
    MobileVLM-3B & MobileLLaMA-2.7B  & 558K & 665K& - & {59.0} & 61.2 &47.5 & 84.9 &  59.6 & -\\
    \midrule
    VL-Mamba& Mamba LLM{-2.8B} &558K & 665K&{76.6} & 56.2 & 65.4 & {48.9} &{84.4} & 57.0& {}\\
    \midrule
    \rowcolor{blue!5}
    \method& RWKV7 LLM{-3B} &558K & 665K&{78.3} & 60.8	 & 70.9 & {51.1} &\textbf{{87.1}} & 66.6& \textbf{38.7}\\
    \bottomrule
    \end{tabular}}
\end{table*}

\begin{table*}[t]
    \small
    \centering
    \caption{Model's WER(\%) on Librispeech dataset and CER(\%) on Aishell-1 dataset.}
    \begin{tabular}{lclcccc}
    \toprule
    \textbf{Dataset} & \textbf{ Data (h)}& \textbf{Encoder} & \textbf{Clean WER(\%)}& \textbf{Other WER(\%)}& \textbf{Dev CER(\%)} & \textbf{Test CER(\%)} \\
    \midrule
    \multirow{4}{*}{Librispeech} & \multirow{4}{*}{960} & wavlm large & 2.43 & 6.51 & - & - \\
     & & wavlm base+ & 3.08 & 10.38 & - & - \\
       & & whisper medium & 5.33 & 12.28 & - & - \\
       & & whisper small & 6.24 & 16.92 & - & - \\
    \midrule
    \multirow{4}{*}{Aishell-1} & \multirow{4}{*}{178} & wavlm large & - & - & 9.68 & 10.33 \\
     & & wavlm base+ & - & - & 12.40 & 13.46 \\
      & & whisper medium & - & - & 5.08 & 5.83 \\
       & & whisper small & - & - & 6.29 & 6.95 \\
    \bottomrule
    \end{tabular}
    \label{audio_performance}
    \vspace{-10pt}
\end{table*}

\section{Experiments}
\subsection{Experimental Details}
\paragraph{Training Settings} (1) \textit{Vision}. Our implementation follows the phased training paradigm of LLaVA \cite{liu2023visualinstructiontuning} for both vision and audio understanding. In Phase I, we first freeze the encoder and the RWKV model, training only a linear adapter with a single MLP and layer norm to project multimodal features into the embedding space of the language model. In Phase II, we then unfreeze both the adapter and RWKV parameters, while the encoder remains frozen to preserve pretrained representations. To comprehensively assess the impact of encoder choice and model scale on RWKV7 performance, we performed experiments on four vision languagemarks using three model sizes (0.4B, 1.5B and 3B) for each encoder. Our models are trained on 8$\times$NVIDIA A800 GPUs. Details of training settings can be found at Appendix~\ref{table:visual-hyper}. (2) \textit{Audio}. Training was conducted in two phases: Phase I trained only the audio adapter (LR=1e-4), while Phase II jointly trained the adapter and RWKV (LR decayed from 1e-4 to 5e-5). For LibriSpeech, we ran 1 epoch in each phase; for Aishell-1, 2 epochs in Phase I and 4 in Phase II. The default batch size was 32, reduced to 16 for the Whisper encoder due to GPU constraints, with epochs halved accordingly to match training steps. All experiments used 44$\times$090 GPUs. (3) \textit{Time series}. In the Time series task, We conducted experiments using dual NVIDIA RTX 4090 (24GB) GPUs, training on a 441,725-sample short-duration univariate dataset.

    


\begin{table*}[t]
    \small
    \centering
    \caption{Zero-shot MSE with Adapter Scaling 4$\times$ use gift-eval datasets (WaveNet Encoder)~\cite{qiu2024tfb}}
    \begin{tabular}{lcccccccc}
    \toprule
    \textbf{Model} &\textbf{LB-FL} &  \textbf{ECL} & \textbf{ETTh1} & \textbf{ETTh2} & \textbf{ETTm1} & \textbf{ETTm2} & \textbf{WTH} & \textbf{Traffic} \\
    \midrule
    TimeFM & 720-96 & \textbf{0.119} & 0.421 & 0.326 & 0.363 & 0.206 & \textbf{0.123} & \textbf{0.327} \\
    Timer & 720-96 & 0.221 & 0.414 & 0.305 & 0.440 & 0.203 & 0.178 & 0.526 \\
    UniTS & 720-96 & 0.175 & 0.377 & 0.323 & 0.761 & 0.249 & 0.194 & 0.481 \\
    TTM & 720-96 & 0.170 & \textbf{0.368} & 0.286 & 0.415 & \textbf{0.186} & 0.152 & 0.509 \\
    MOIRAI & 720-96 & 0.212 & 0.394 & \textbf{0.285} & 0.516 & 0.222 & 0.208 & 1.359 \\
    ROSE & 720-96 & 0.209 & 0.382 & 0.298 & 0.512 & 0.224 & 0.200 & 0.572 \\
    \midrule
    \rowcolor{blue!5}
    \textbf{\method (25\% gift-eval)} & 720-96 & 0.342 & 0.746 & 0.633 & 0.754 & 0.559 & 0.797 & 0.512 \\
    \rowcolor{blue!5}
    \textbf{\method (100\% gift-eval)} & 720-96 & 0.342 & 0.648 & 0.453 & \textbf{0.227} & 0.426 & 0.203 & 0.342 \\
    \bottomrule
    \end{tabular}
    \label{tab:time_encoder_bench}
    \vspace{-10pt}
\end{table*}

\paragraph{Datasets}
We consider diverse datasets in vision, audio, and time series (Refer to Table~\ref{tab:multimodal_benchmarks}). For vision understanding ability, we use LLaVA-595K as training dataset for Phase I, and LLaVa-665k for Phase II. For audio, We train our \method model using two open-source datasets: (1) LibriSpeech \cite{7178964}, which comprises 960 hours of English reading audio data; and (2) Aishell-1 \cite{bu2017aishell1opensourcemandarinspeech}, which includes 170 hours of Chinese audio data. For each, we trained our model exclusively on the respective training dataset. In the time series task, we utilized public datasets from GIFT-Eval \cite{aksu2024giftevalbenchmarkgeneraltime}. After thorough sorting and cleaning, we derived a small number of univariate datasets. Additionally, we incorporated UTSD \cite{liu2024timer} public datasets later in the process. 

\paragraph{Benchmarks}
To rigorously evaluate our model's capabilities across diverse reasoning scenarios, we employed a comprehensive evaluation framework spanning from basic visual recognition to advanced knowledge-intensive tasks. This framework systematically verifies our model's cross-modal competence at various cognitive levels by assessing it on seven multimodal benchmarks: VQA-v2~\cite{goyal2017vqav2} for fundamental image understanding and question-answering, TextVQA~\cite{singh2019textvqa} to evaluate optical character recognition (OCR) and text-image integration, GQA~\cite{hudson2019gqa} for compositional reasoning and real-world visual understanding, ScienceQA~\cite{lu2022learn} to assess scientific multimodal reasoning through multiple-choice questions, POPE~\cite{li2023pope} to quantify object hallucination via binary classification tasks, MMMU~\cite{yue2024mmmumassivemultidisciplinemultimodal} to challenge models with college-level, cross-discipline problems, and MMBench~\cite{liu2024mmbenchmultimodalmodelallaround}, which represents a systematically designed, objective evaluation framework for comprehensive assessment that uses circularEval strategy for assessment stability, ETT~\cite{qiu2024tfb}, which focuses on long-term multivariate time-series forecasting using electricity transformer temperature data, serving as a standard benchmark for evaluating temporal modeling capabilities under various sequence lengths and prediction horizons, WeatherBench~\cite{rasp2020weatherbench} to evaluate spatiotemporal forecasting using global atmospheric data as a standard benchmark for data-driven weather prediction, etc.. Additionally, we evaluated our \method model using the corresponding open-source datasets: LibriSpeech~\cite{7178964}, which comprises 960 hours of English reading audio data, and Aishell-1~\cite{bu2017aishell1opensourcemandarinspeech}, which includes 170 hours of Chinese audio data.

\subsection{Qualitative Evaluation}

\paragraph{Vision Understanding}

As summarized in Table~\ref{tab:results}, \method \ demonstrates strong overall performance across eight widely-used multimodal benchmarks, outperforming existing state-of-the-art (SoTA) methods in its parameter range. Compared to VL-Mamba-2.8B, \method-3B consistently achieves higher scores on all evaluated tasks, reflecting its superior capability in visual question answering, compositional reasoning, and image-conditioned instruction following.

Notably, despite having a significantly smaller language backbone than LLaVA-1.5-7B, \method \ achieves competitive or superior results on several benchmarks. It surpasses LLaVA-1.5-7B in ScienceQA-IMG, POPE, and MMBench, while maintaining comparable performance on VQAv2. Furthermore, \method \ attains the highest reported score among peers on the MMMU benchmark, highlighting its generalization ability in challenging multi-modal understanding scenarios.

These results collectively suggest that \method \ offers a favorable trade-off between performance and model size. Its effectiveness stems not merely from scale, but from architectural efficiency and a well-designed multimodal integration strategy, positioning it as a competitive alternative to larger vision-language models.

\begin{table*}[t!]
    \small
    \centering
    \caption{Zero-shot MSE on Public Datasets: ECL, ETTh, ETTm, WTH, Traffic (Timer Encoder)~\cite{liu2024timer}}
    \begin{tabular}{lcccccccc}
    \toprule
    \textbf{Dataset Size} &\textbf{Adapter Scaling} &  \textbf{ECL} & \textbf{ETTh1} & \textbf{ETTh2} & \textbf{ETTm1} & \textbf{ETTm2} & \textbf{WTH} & \textbf{Traffic} \\
    \midrule
    Gift-Evel & 2$\times$ & 0.641 & 0.785 & 0.882 & 0.949 & 0.719 & 0.633 & 0.988 \\
    Gift-Evel + UTSD & 2$\times$ & 0.516 & 0.637 & 0.848 & 0.891 & 0.672 & 0.512 & 0.683 \\
    Gift-Evel + UTSD & 4$\times$ & \textbf{0.453} & \textbf{0.629} & \textbf{0.547} & 0.843 & \textbf{0.648} & \textbf{0.461} & 0.641 \\
    Gift-Evel + UTSD & 8$\times$ & 0.535 & 0.629 & 0.652 & \textbf{0.828} & 0.762 & 0.566 & \textbf{0.617} \\
    \bottomrule
    \end{tabular}
    \label{tab:time_series_scale_bench}
\end{table*}

\begin{table}[t]
    \small
    \centering
    \caption{\method \ Visual Models with different Encoders and parameters tested on benchmarks.}
    \begin{tabular}{lcccccc}
        \toprule
        \textbf{Vision} & \textbf{ Size} & \textbf{VQA}$^\text{v2}$& \textbf{VQA}$^\text{T}$  & \textbf{GQA} & \textbf{SQA}$^\text{I}$ \\
        \midrule
        \multirow{3}{*}{CLIP} & 0.4B & 62.04& 31.72& 49.32& 51.10 \\
         &1.5B& 72.31& 40.27& 54.56& 62.77\\
           & 3B& 73.13& 45.56& 57.00& 70.66  \\
           
        \midrule
        \multirow{3}{*}{SigLIP2} &0.4B& 72.04 & 38.75& 55.52& 43.32 \\
         &  1.5B& 76.95 & 44.96& 58.88& 63.10 \\
          &  3B& 78.30 & 51.09& 60.75& 70.93 \\
        
        \bottomrule
    \end{tabular}
    \label{visual_performance}
\end{table}

\paragraph{Vision Knowledge}
The following examples in Table~\ref{tab:visual_example} showcase the capabilities of the \method \ QA chatbot. These examples illustrate how \method \ effectively integrates visual information with general knowledge, while also performing basic logical reasoning to address common user queries.

\paragraph{Audio Recognition}
Table \ref{audio_performance} presents the Word Error Rate (WER) for the LibriSpeech test\_clean and test\_other test sets, as well as the Character Error Rate (CER) for the Aishell-1 development and test sets. For the LibriSpeech dataset, the model achieved a WER of 2.43\% on the test\_clean subset, indicating a high level of precision in recognizing clear speech. On the test\_other subset, the model attained a WER of 6.51\%, demonstrating reasonable performance in handling more challenging noisy speech samples without data augmentation. For the Aishell-1 dataset, the model achieved CERs of 5.08\% on the development set and 5.83\% on the test set, using the Whisper medium encoder. These results reflect the model's effectiveness in handling non-English speech recognition tasks with limited training data.

During adapter training, we observed a phenomenon akin to the \textit{capability emergence} described by \cite{ma2024embarrassinglysimpleapproachllm}. However, the timing of this emergence was inconsistent and heavily influenced by the initialization of the adapter's weights. In some instances, the adapter failed to converge during Phase I.
\paragraph{Time Series Forecasting}


We conducted comparative experiments on two temporal encoder architectures: Timer and WaveNet. Results (See Table~\ref{tab:time_encoder_bench}) show that although Timer has a larger parameter count (based on pre-trained weights), it consistently underperforms WaveNet on downstream time-series forecasting tasks. We hypothesize that this performance gap arises from WaveNet’s use of causal dilated convolutions, which effectively capture long-range temporal dependencies through hierarchically expanding receptive fields. Additionally, unlike Timer’s patch-wise embedding, WaveNet adopts a point-wise embedding strategy, allowing it to extract finer-grained temporal features. For training data preparation, we constructed two fine-tuning datasets: a baseline dataset (GIFT-Eval)\cite{aksu2024giftevalbenchmarkgeneraltime} and an augmented dataset composed of GIFT-Eval and a partially processed subset of UTSD\cite{liu2024timer}. Experiments indicate that models trained on the augmented dataset—containing anomalous samples—achieved superior generalization in zero-shot evaluations across public benchmarks including ECL, ETT, WTH, and Traffic. Notably, this training strategy enables the model to maintain stable predictions even under distribution shifts, demonstrating strong robustness and generalization. Architecture ablation studies further revealed that the scaling factor of the adapter modules plays a significant role in performance. A scaling factor of 4$\times$ yielded the best overall results on the validation set (see Table \ref{tab:time_series_scale_bench}), outperforming the 8$\times$ and 2$\times$ settings by approximately 10.0\% and 13.5\%, respectively.\\

 Overall, even under constrained conditions—no data augmentation, limited training data, and fewer training steps—the \method \ model achieved competitive accuracy on time-series forecasting tasks, providing empirical evidence for its applicability in real-world, complex scenarios.

\subsection{Ablation Study}
\paragraph{The Effect of Different Vision Encoders}
In order to evaluate the impact of different vision encoders on the performance of multimodal models, this study designed rigorous comparative experiments. We selected two representative visual encoder architectures for comparison: the contrastive learning-based CLIP and the recently proposed SigLIP2. In the experimental design, we specifically controlled the following variables: the length of the encoded visual feature sequences for both (google/siglip2-base-patch16-384 and openai/clip-vit-large-patch14-336) was set to 577, to eliminate any potential confusion caused by differences in sequence length that could affect the understanding ability of LLMs; cross-validation was conducted on LLMs of different scales (ranging from 0.4B to 3B parameters) to ensure the generalizability of the experimental conclusions. 

\begin{table}[t]
    \small
    \centering
    \setlength{\tabcolsep}{5pt}
    \caption{By controlling the kernel and stride of conv1d, control the sequence length of multimodal signals to compare performance differences.}
    \begin{tabular}{lcccccc}
    \toprule
    \textbf{Size} & \textbf{(k,s)}& \textbf{ Token} & \textbf{VQA}$^\text{v2}$& \textbf{VQA}$^\text{T}$  & \textbf{GQA} & \textbf{SQA}$^\text{I}$ \\
    \midrule
    \multirow{4}{*}{1.5B} & (0,0)& 577 &76.95 & 44.96& 58.88& 63.10 \\
    & (3,2)&288& 75.21& 45.75& 58.28& 66.02\\
    & (4,3)&192&74.17 & 44.27& 57.53& 65.72 \\
    & (5,4)&144&73.21 & 42.65& 57.07& 65.29 \\
    \bottomrule
    \end{tabular}
    \label{tab:conv_performance}
    \vspace{-10pt}
\end{table}

As shown in Table~\ref{visual_performance}, SigLIP2 encoder consistently outperforms CLIP encoder all evaluated benchmarks, including VQAv2, TextVQA, GQA, and ScienceQA. Notably, the SigLIP2-based model achieves significant improvements in both general and text-based visual question answering tasks, as well as in compositional reasoning. Despite its encoder containing only 90M parameters—approximately 30\% of the CLIP encoder’s size—SigLIP2 demonstrates superior performance, particularly in tasks requiring fine-grained visual-text alignment and semantic understanding. These results underscore that model effectiveness in multimodal understanding is influenced more by encoder design and pretraining methodology than by parameter scale alone.

\paragraph{Efficiency of Sequence Compression via 1D Convolution}

It is well known that the efficiency problem in processing long sequences has long been one of the main bottlenecks limiting the performance of LLMs. This challenge is particularly prominent in multimodal tasks, where signals from different modalities often generate a large number of tokens after encoding. For example, in the \method model, a single image encoded through the SigLIP2 encoder generates 577 tokens, and when extended to video sequences, the length increases by an order of magnitude. To address this issue, this section systematically investigates the optimization effects of convolutional dimensionality reduction (Conv1D), aiming to provide new technical insights for sequence compression research.

We conducted empirical research (See Table~\ref{tab:conv_performance} and visualization in Figure~\ref{fig:design}) on the \method-1.5B model architecture using the LLaVA training dataset, and performed comprehensive evaluations across multiple benchmark datasets, including VQAv2, TextVQA, GQA, and ScienceQA. The experimental results show that when the sequence length is compressed by 50\%, the model exhibits only a slight decrease in performance (on average) while achieving a 4.6\% accuracy improvement on the ScienceQA task. Further research reveals that as the kernel size and stride increase, although the model performance exhibits a gradual decline, the computational efficiency is significantly improved. We tested \method-1.5B on single 4090 GPU without any acceleration; The results indicate that increasing the compression ratio of token sequences can substantially accelerate inference speed, showing a clear efficiency gain. This highlights an effective strategy for balancing computational efficiency and model performance, offering valuable insights for practical deployment.

\paragraph{G1 reasoning model} We experimentally validate the effect of text pretraining weights on the ability of large language models to understand multimodal information by comparing two pretraining weights (base and g1) of the RWKV7-0.4B model. It is important to note that the g1 model is an improved version of the base model, obtained through post-training by introducing a large amount of 'think'- type data. Although both models perform similarly in pure text NLP benchmark tests, as shown in Table~\ref{table:g1}, fine-tuning with the g1 pretraining weights significantly outperforms the base model across all metrics, with an exceptionally significant improvement observed in the SQA metric (specific improvement is 28\%). This empirical result strongly confirms that an appropriate text pretraining strategy can effectively enhance the language model's ability to understand multimodal information, thereby improving its overall performance in downstream tasks.

\begin{figure*}[ht]
    \centering
    \includegraphics[width=0.79\textwidth]{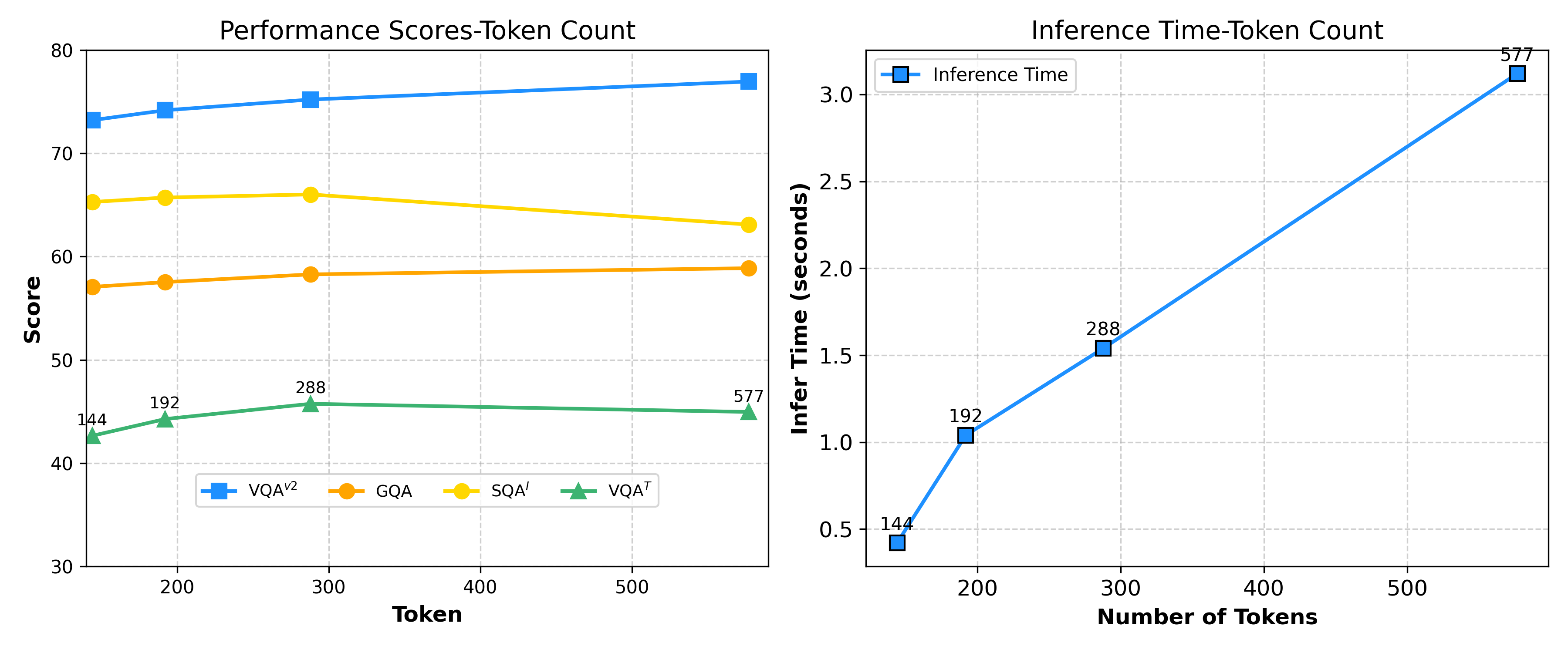}
    \caption{Performance and efficiency of \method. \textbf{Left.} The scaling curve of tokens with the performance score. \textbf{Right.} The inference time of \method \ with the number of tokens.}
    \label{fig:design}
    \vspace{-15pt}
\end{figure*}

\begin{table}[t]
\small
\centering
\caption{Performance differences under different pretraining weights}
\begin{tabular}{lcccccc}
\toprule
\textbf{Size} & \textbf{ Model} & \textbf{VQA}$^\text{v2}$& \textbf{VQA}$^\text{T}$  & \textbf{GQA} & \textbf{SQA}$^\text{I}$ \\

\midrule
\multirow{2}{*}{0.4B} &base& 72.04 & 38.75& 55.52& 43.32 \\
 &  g1& 73.21 & 41.13& 57.34& 55.58 \\
\midrule
\multirow{2}{*}{1.5B} &base& 76.95 & 44.96& 58.88& 63.10 \\
 &  g1& 77.87 & 50.91& 60.18& 64.63 \\

\bottomrule
\end{tabular}
\label{table:g1}
\vspace{-10pt}
\end{table}

\paragraph{Time Series Forecasting encoder Compare with Timer and WaveNet}
In a feedforward neural network (FFN), activation functions such as ReLU introduce sparsity by setting some outputs to zero, which in turn reduces the rank of the output matrix and may impact the model’s representational capacity. Through both theoretical analysis and empirical experiments using the results in Table \ref{tab:time_series_scale_bench}, we observed that the effect is suboptimal when the hidden layer dimension is set to 2x or 8x the input dimension.

Table \ref{tab:time_series_scale_bench} presents the zero-shot mean squared error (MSE) performance of different adapter scaling configurations on multiple public datasets, including ECL, ETTh, ETTm, WTH, and Traffic. The results indicate that increasing the adapter scaling factor from 2$\times$ to 4$\times$ significantly improves performance across most datasets, with the lowest MSE values observed at 4$\times$ scaling. Specifically, the Gift-Evel + UTSD model with 4$\times$ scaling achieves the best results on ECL (0.453), ETTh1 (0.629), ETTh2 (0.547), ETTm2 (0.648), WTH (0.461), and Traffic (0.641), demonstrating that this configuration effectively enhances model accuracy. 

However, further increasing the scaling factor to 8$\times$ does not consistently improve performance, with some datasets showing increased error values. This suggests that excessively large hidden layer dimensions may introduce instability or diminish representational efficiency. Based on these findings, we recommend setting the hidden layer dimension to at least four times the input dimension to preserve sufficient rank, thereby enhancing the model’s representational power and stability.
\[
p = 1 - \frac{\sum_{i=m}^{n} \binom{n}{i}}{2^n}
\]
Table \ref{tab:time_encoder_bench} presents the zero-shot mean squared error (MSE) results for various models using the WaveNet encoder with adapter scaling 4$\times$ on public datasets, leveraging the gift-eval dataset. The models are evaluated on multiple time-series forecasting benchmarks, including ECL, ETTh, ETTm, WTH, and Traffic, with a lookback length of 720 and a forecast length of 96.

From the results, {TimeFM} achieves the best performance on ECL (0.119), WTH (0.123), and Traffic (0.327), demonstrating strong predictive capabilities on these datasets. {TTM} performs best on ETTh1 (0.368) and ETTm2 (0.186), while {MOIRAI} achieves the lowest error on ETTh2 (0.285). Our proposed model, {\method} (100\% gift-eval), outperforms other models on ETTm1 (0.227), showing its effectiveness in short-term forecasting for this dataset.

Comparing \method (25\% gift-eval) and \method (100\% gift-eval), we observe that increasing the proportion of gift-eval data significantly improves performance across most datasets, particularly on ETTh2 (from 0.633 to 0.453) and ETTm1 (from 0.754 to 0.227). This suggests that leveraging a larger portion of the gift-eval dataset enhances ours model generalization and stability.

Overall, the results highlight the varying strengths of different models across datasets, emphasizing the importance of dataset composition and model architecture in achieving optimal forecasting performance.
\section{Conclusion}

In this paper, we propose \method, a multimodal understanding framework that enables modality switching via interchangeable encoders. Built upon RWKV7, \method \ provides a comprehensive analysis and evaluation of the capabilities of modern RNN architectures in the multimodal domain.

\section{Limitations}
This paper presents a systematic evaluation of the proposed \method \ framework across a range of benchmark tasks involving different modalities, demonstrating the feasibility of applying linear-structured models to multi-modal large language models (MLLMs). 
Nonetheless, this work does not yet explore more complex multi-modal fusion scenarios, such as tri-modal tasks involving speech, vision, and language. Future work will aim to address these richer multi-modal settings.

\bibliography{custom}
\bibliographystyle{acl_natbib}

\newpage
\appendix
\section{Example Appendix}
\begin{table}[h]
    \centering
    \small
    \caption{Hyperparameter Settings for Visual Instruction Tuning of \method
 }
    \begin{tabular}{lll}
    \hline
    \textbf{Hyperparameter} & \textbf{Step1} & \textbf{Step2} \\ \hline
    
    bsz & 256 & 128 \\
    ctx\_len & \multicolumn{2}{c}{2048} \\
    lr & 1e-3 & 2e-5 \\
    warmup\_steps & \multicolumn{2}{c}{ 100} \\
    beta1 & \multicolumn{2}{c}{ 0.9} \\
    beta2 & \multicolumn{2}{c}{ 0.99} \\
    adam\_eps & \multicolumn{2}{c}{ 1e-8} \\
    precision & \multicolumn{2}{c}{ bf16} \\
    lr\_schedule & \multicolumn{2}{c}{wsd} \\ \hline
    
    \label{table:visual-hyper}
    \end{tabular}
\end{table}

\subsection{Examples}

\begin{table*}[htbb]

\small
\begin{minipage}{0.99\textwidth}

\centering  
\vspace{-4mm}
\scalebox{0.88}{
\begin{tabular}{l p{12.5cm} }
\toprule

 \multicolumn{2}{l}{\bf Visual Examples}  \\
\midrule
&  \includegraphics[height=4.5cm]{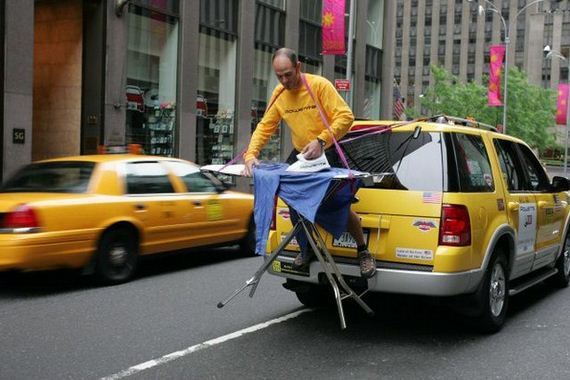} \\

User & How many Cars in this picture? \\
\method & There are two cars in the picture, with one being the taxi cab the man is standing on.\\

\midrule
 & [{\it Start a new conversation, and clear the history}] \\
 &  \includegraphics[height=4.5cm]{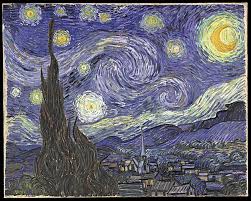}
\\ \midrule 
User & What is the name of this picture?  \\
\method & The name of this picture is “Starry Night ”\\

User & Who drew it? \\
\method & \textcolor{red}{Vincent van Gogh drew this picture.} \\

\midrule  
 & [{\it Start a new conversation, and clear the history}] \\
 &  \includegraphics[height=4.5cm]{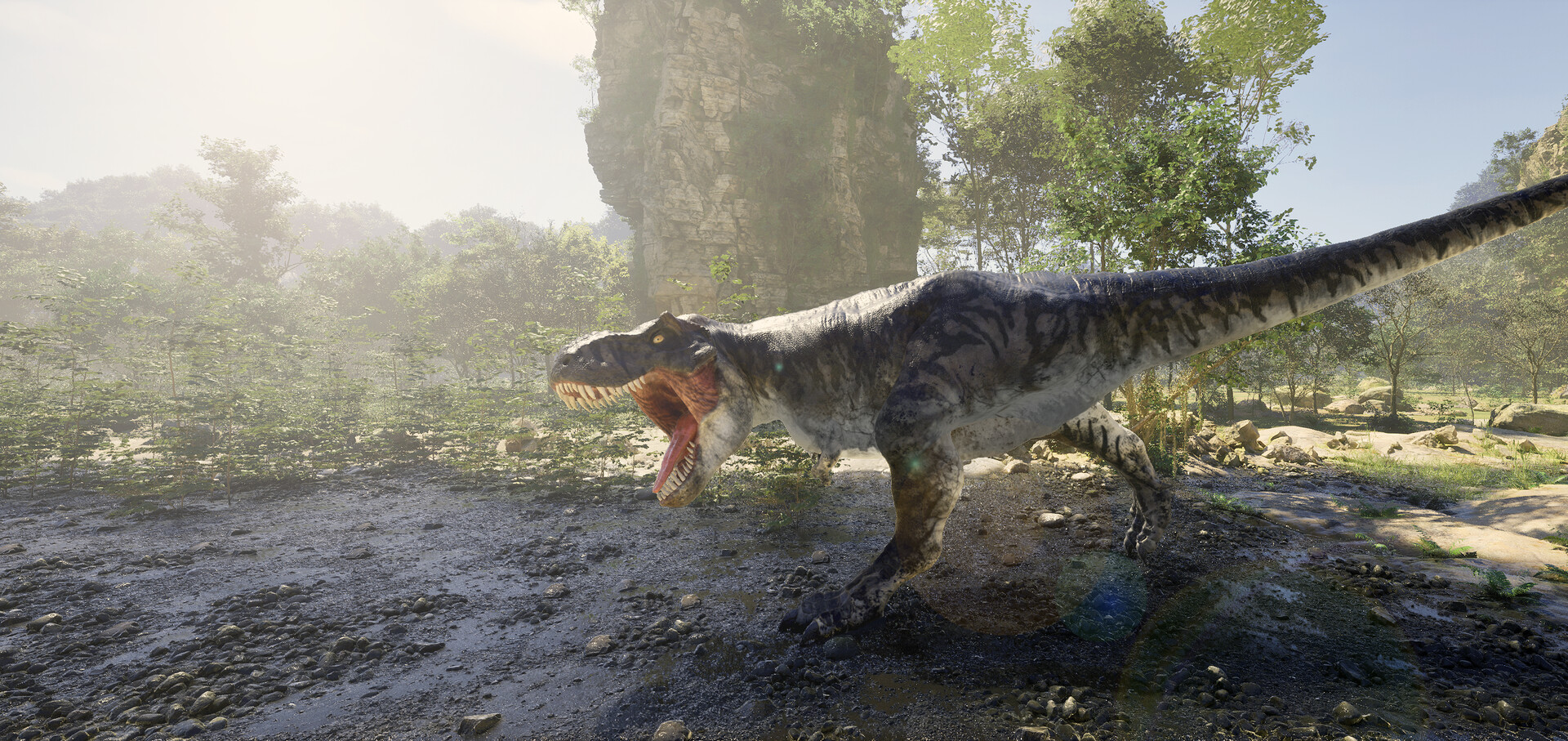}
\\ \midrule
User & Is this real in current life? \\
\method & No, this is not a real animal. \textcolor{blue}{It is a computer-generated image (CGI) of a dinosaur, specifically a T-Rex, walking through a forest.}
\\
User & Depict the environment \\
\method & The environment in the image is a forest with trees, rocks, and a dirt path. The CGI dinosaur is walking through this natural setting.\\
User & Based on the facial expression, what is it's emotion?  \\
\method & \textcolor{blue}{The dinosaur appears to be angry, as it is depicted with its mouth open and teeth showing.}\\
User & Based on its emotion and the environment, what do you think this dinosaur is doing right now?  \\
\method & Based on its angry expression and the forest setting, it is possible that the dinosaur is either searching for food, defending its territory, or reacting to a perceived threat. \textcolor{blue}{The fact that it is walking through the forest suggests that it might be exploring its habitat or looking for prey}. However, without more context or information, it is difficult to determine the exact reason for its current behavior.\\
\bottomrule
\end{tabular}

}
\captionof{table}{Visual Example.}
\label{tab:visual_example}  
  \end{minipage}
\end{table*}

\begin{table*}[htbb]
\begin{minipage}{0.99\textwidth}
\centering  
\vspace{-4mm}
\scalebox{0.88}{
\begin{tabular}{l p{12.5cm}}
\toprule

\multicolumn{2}{l}{\bfseries Audio Examples}  \\
\midrule

User & \raisebox{-0.5ex}{\includegraphics[height=12pt]{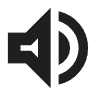}} What is your name? \\

\method & \raisebox{-0.5ex}{\includegraphics[height=12pt]{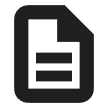}} 
\parbox[t]{11cm}{
\textbf{\large Hello!} My name is RWKV, and I'm your friendly voice assistant here to help with any questions or tasks you have.
} \\

\midrule

User & \raisebox{-0.5ex}{\includegraphics[height=12pt]{Figures/audio_icon.png}} Can I drink coffee before I go to bed? \\

\method & \raisebox{-0.5ex}{\includegraphics[height=12pt]{Figures/text_icon.png}} 
\parbox[t]{11cm}{
Yes, you can drink coffee before you go to bed, but it's important to do so in moderation. \textcolor{red}{Drinking coffee too close to bedtime can interfere with your sleep quality. Try to have your last cup of coffee at least 6 hours before bedtime. If you're sensitive to caffeine, you might want to avoid it altogether. Also, consider other relaxing activities before bed to help you wind down}
} \\

\midrule

User & \raisebox{-0.5ex}{\includegraphics[height=12pt]{Figures/audio_icon.png}} Tell me something about machine learning. \\

\method & \raisebox{-0.5ex}{\includegraphics[height=12pt]{Figures/text_icon.png}} 
\parbox[t]{11cm}{
Machine learning is a branch of artificial intelligence that focuses on creating systems that can learn from and make decisions based on data. \textcolor{blue}{It involves algorithms that enable computers to identify patterns and make predictions or decisions without being explicitly programmed for each task. This technology is widely used in various applications, such as recommendation systems, speech recognition, and autonomous vehicles. }Machine learning models can improve over time as more data is processed, making them more accurate and efficient. It's a powerful tool for automating tasks, enhancing user experiences, and driving innovation across many industries.
} \\

\bottomrule
\end{tabular}
}
\captionof{table}{Interactive Audio Examples Demonstrating Question-Answer Capabilities}
\label{tab:audio_example}  
\end{minipage}
\end{table*}

\end{document}